\documentclass{bmvc2k}

\title{CU-Net: Coupled U-Nets}
\addauthor{Zhiqiang Tang}{zt53@cs.rutgers.edu}{1}
\addauthor{Xi Peng}{xipeng@binghamton.edu}{2}
\addauthor{Shijie Geng}{sg1309@rutgers.edu}{1}
\addauthor{Yizhe Zhu}{yizhe.zhu@rutgers.edu}{1}
\addauthor{Dimitris N. Metaxas}{dnm@cs.rutgers.edu}{1}

\addinstitution{
 Department of Computer Science\\
 Rutgers University\\
 New Jersey, USA
}
\addinstitution{
 Department of Computer Science\\
 Binghamton University\\
 New York, USA
}

\runninghead{Zhiqiang Tang}{Coupled U-Nets}

\usepackage{times}
\usepackage{xcolor}
\usepackage{soul}
\usepackage{booktabs}
\usepackage{adjustbox}
\usepackage{epsfig}
\usepackage{graphicx}
\usepackage{subfigure}
\usepackage{amsmath}
\usepackage{amssymb}
\usepackage{microtype}
\usepackage{multirow}
\usepackage{enumitem}
\usepackage{eso-pic}
\usepackage{bm}
\usepackage{amsmath}

\begin{document}

\maketitle
\begin{abstract}

We design a new connectivity pattern for the U-Net architecture. Given several stacked U-Nets, we couple each U-Net pair through the connections of their semantic blocks, resulting in the coupled U-Nets (CU-Net). The coupling connections could make the information flow more efficiently across U-Nets. The feature reuse across U-Nets makes each U-Net very parameter efficient. We evaluate the coupled U-Nets on two benchmark datasets of human pose estimation. Both the accuracy and model parameter number are compared. The CU-Net obtains comparable accuracy as state-of-the-art methods. However, it only has at least 60\% fewer parameters than other approaches.

\end{abstract}
\section{Introduction}


The U-Net \cite{ronneberger2015u} architecture has been widely used in the location-sensitive tasks such as human pose estimation \cite{newell2016stacked}, semantic segmentation \cite{long2015fully}, etc. The top-down and bottom-up processing facilitates the inference at multiple scales. The shortcut connections between the corresponding top-down and bottom-up blocks help keep the spatial information.

More recently, the DenseNet \cite{huang2016densely} has shown superior performance in both image classification accuracy and parameter efficiency than the ResNet \cite{he2016deep}. The dense connectivity improves the feature reuse in the network forward process and the gradient propagation during the backward process. Thus, it could use less parameters to achieve comparable or even better accuracy. A natural question arises: how could we use the dense connectivity to improve the performance of the U-Net?

Some works \cite{jegou2017one, li2017h} tried to combine the dense connectivity and the U-Net. They follow the DenseNet design. In particular, each top-down or bottom-up resolution has a dense block containing several densely connected convolutional layers. This straightforward application of dense connectivity is restricted within local blocks of a single U-Net. Another question arises: could we integrate the dense connectivity into several stacked U-Nets?

\begin{figure*}[t!]
\centering
  \includegraphics[width=1.0\linewidth]{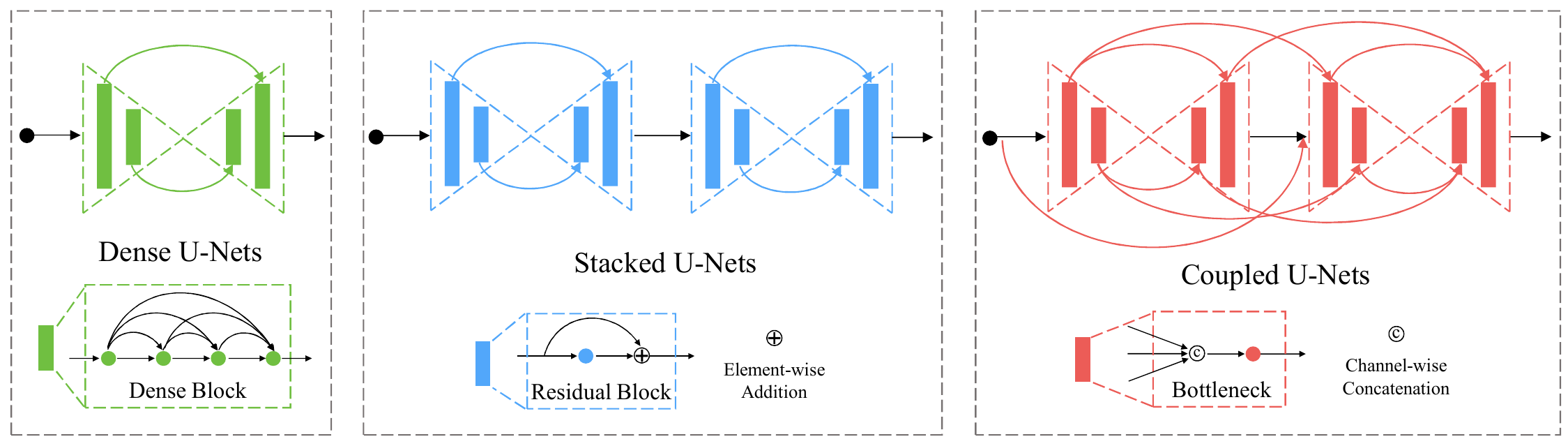}
\caption{Illustration of naive dense U-Net, stacked U-Nets and coupled U-Nets (CU-Net). The dense U-Net and stacked U-Nets have shortcut connections only inside each U-Net. In contrast, coupled U-Nets also have connections for semantic blocks across U-Nets. The CU-Net is a hybrid of dense U-Net and stacked U-Net, integrating the merits of both dense connectivity and multi-stage top-down and bottom-up refinement.}
\label{fig:framework}
\end{figure*}

In this paper, we propose a global connection pattern. Given several stacked U-Nets, we add shortcut connections for each U-Net pair, generating the coupled U-Nets (CU-Net). The key idea is we connect blocks of the same semantic meanings, {\it i.e.} having the same resolution in either top-down or bottom-up context. Please refer to Figure \ref{fig:framework} for an illustration. Basically, a pair of U-Nets are connected at both top-down and bottom-up context.


The proposed coupled U-Nets have three merits. {\bf First}, the coupling connections are global, extending from the first U-Net to the last one. It encourages the feature reuse as well as gradient propagation globally across different U-Nets. In contrast, the straightforward application of dense connectivity only helps the information flow inside a single U-Net. {\bf Second}, we could easily add a supervision at the end of each U-Net if several U-Nets are coupled together. In other words, the coupled U-Nets could naturally take advantage of multiple supervisions. However, a single dense U-Net generally only has one supervision at the end. {\bf Third}, the coupled U-Nets also preserve the advantage of stacked U-Nets. Generally, several stacked U-Nets could achieve higher accuracy than a large U-Net of the equivalent model size. This benefits from the multi-stage top-down and bottom-up inference along the U-Net cascade. The proposed coupled U-Nets still inherit this nice property. Furthermore, the U-Nets coupling could largely improve the information flow based on the traditional stacked U-Nets. This could significantly reduce the model parameter number, yielding very compact models. In summary, our key contributions are:

\begin{itemize}
    \item To the best of our knowledge, we are the first to propose coupled U-Nets (CU-Net) by connecting semantic blocks of pairwise U-Nets. The information can flow more efficiently and the feature reuse across U-Net pairs makes each U-Net light-weighted.
    \item We investigate to use intermediate supervisions with coupled U-Nets. With a moderate amount of intermediate supervisions, the coupled U-Nets could get the highest accuracy. We also observe that full intermediate supervisions are not the optimal choice.
    \item Exhaustive experiments are conducted on the human pose estimation. CU-Net demonstrates superior localization accuracy and use at least 60\% less parameters compared with state-of-the-art methods.
\end{itemize}



\section{Related Work}
In this section, we review recent work on designing convolutional network architectures and recent developments in human pose estimation.

{\bf Network Architecture.}
The research on network architectures have been active since
AlexNet \cite{alex2012alexnet} appeared. First, by using smaller filters, the VGG \cite{simonyan2014very} network become several times deeper than the AlexNet and obtain much better performance. Then the Highway Networks \cite{srivastava2015training} could extend its depths to more than 100 layers with the shortcut connections. Furthermore, the identity mappings make it possible to train ResNet \cite{he2016deep} with more than one thousand layers. More recently, the DenseNet \cite{huang2016densely} outperforms the ResNet benefitting from its dense connections.

The U-Net \cite{ronneberger2015u} architecture was proposed for the biomedical image segmentation. It has been used in semantic segmentation \cite{long2015fully}, face alignment  \cite{peng2016recurrent}, etc. Newell {\ et al.} \cite{newell2016stacked} use the stacked U-Nets in human pose estimation. The also apply the residual module \cite{he2016deep} in the stacked U-Nets. Recently, some efforts \cite{jegou2017one, li2017h} try to bring the dense connectivity \cite{huang2016densely} into the U-Net. However, their shortcut connections are only within a single U-Net.


{\bf Human Pose Estimation.}
CNNs based approaches \cite{wei2016convolutional,pishchulin2016deepcut,lifshitz2016human,zhao2018learning} dominate the human pose estimation and prediction. Newell {\ et al.} \cite{newell2016stacked} apply the stacked U-Nets and get high estimation accuracy. Nearly all recent state-of-the-art methods \cite{chu2017multi,yang2017learning,yu2017adversarial,peng2018jointly} build on it. They use more sophisticated modules, graphical models, or additional adversarial networks \cite{tian2018cr,zhu2018generative}. We focus on largely reducing the model parameters but still obtaining comparable accuracy. 

\section{Network Architecture}
In this section, we first introduce a naive dense U-Net and recap the stacked U-Nets. After analyzing their strengths and weaknesses, we propose a new architecture coupled U-Nets. We also discuss using coupled U-Nets with intermediate supervisions.


\subsection{Naive Dense U-Net}
A U-Net \cite{ronneberger2015u} contains the same number of top-down and bottom-up blocks. There are usually skip connections between them. An illustration is shown in Figure \ref{fig:framework}. The main difference the naive dense U-Net from the traditional U-Net is the previous convolution layers become dense blocks. More specifically, the successive convolution layers at the same spatial resolution are densely connected, forming a dense block. 

Besides, the dense connections result in increasing feature channels in the dense block. To adapt the feature channel number, the 1$\times$1 convolution is used to after each dense block to compress the features.


The dense connections could increase the information flow in the U-Net to some extent. However, they are only within the local blocks. Besides, The naive dense U-Net has only one single U-Net. If we have several U-Nets, is there any more specific design?

\begin{figure*}[t!]
\centering
  \includegraphics[width=\linewidth]{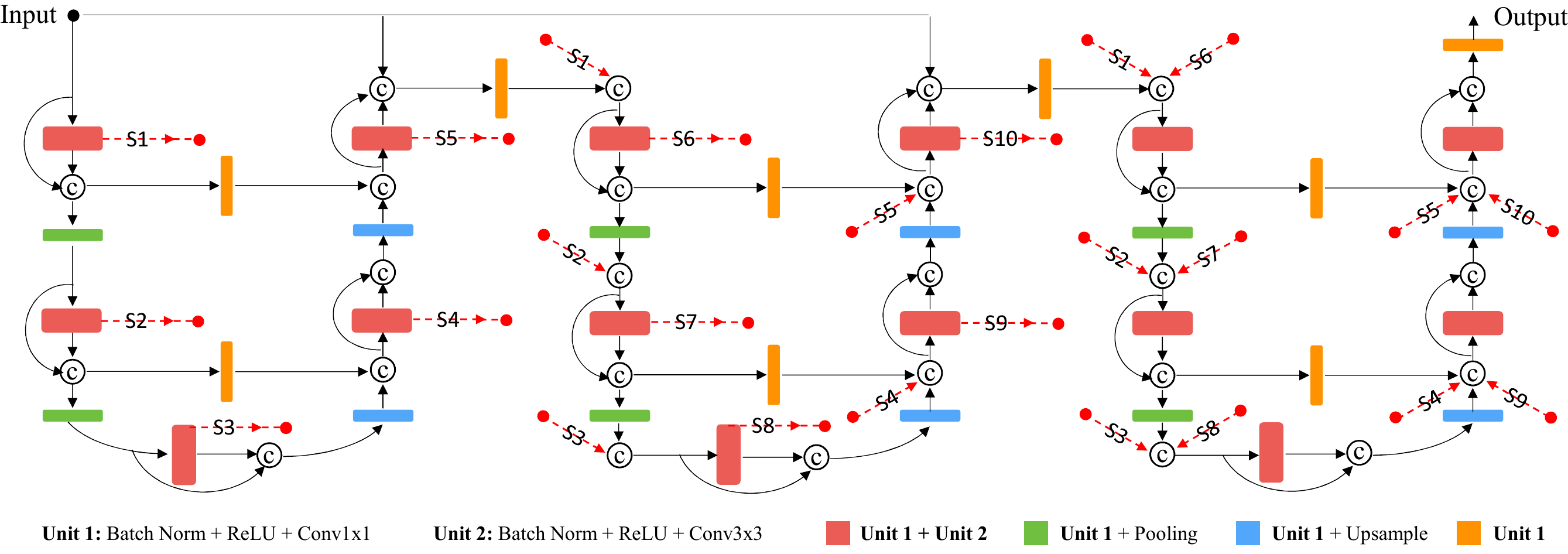}
\caption{Diagram of coupled U-Nets. 3 U-Nets are coupled together by the red dot lines. The red dot lines with the same labels are connected. The same semantic blocks in different U-Nets are connected directly. For simplicity, we only show 2 top-down and bottom-up semantic blocks in each U-Net. The connectivity is similar for more semantic blocks.}
\label{fig:coupled-unet}
\end{figure*}
\subsection{Stacked U-Nets}
Recently, some works \cite{newell2016stacked,wei2016convolutional}  stack multiple U-Nets together. Figure \ref{fig:framework} gives an illustration of stacked U-Nets. Basically, the features would go sequentially from the first U-Net to the last one. The last U-Net makes the final prediction of the model. 

An advantage of stacked U-Nets is its repeated top-down and bottom-up inference. In one U-Net, the input goes through the top-down and bottom-up pipeline once. The U-Net could capture some spatial relationships of the predictions. However, sometimes they may be not enough for the accurate predictions. For instance, in human pose estimation, the relations of upper and lower body joints are complex. Adding a U-Net on top of another could help capture higher order spatial relationships, resulting in higher prediction accuracy.

Besides, the stacked U-Nets make it very easy to add intermediate supervisions. Each U-Net could extend a side path to make its own prediction. We could the same groundtruth for each prediction. It does not affect the feature flow in the main U-Nets cascade. However, it is not straightforward to use intermediate supervisions in a single U-Net. The intermediate supervisions on its top-down blocks encourage predictions ignorant of global cues in the lower resolutions. Similarly, the intermediate supervisions in its bottom-up blocks cannot evaluate the feature effectiveness in the higher resolutions. 

Since stacked U-Nets have more advantages than a single U-Net, could we incorporate the dense connectivity into them? The hybrid should keep the merits of both stacked U-Nets and dense connectivity.

\subsection{Coupled U-Nets}
Although U-Nets stacked together could refine the prediction stage-by-stage, there is no communication among them except for their inputs and outputs. To make information flow more efficiently across different U-Nets, we propose to couple U-Net pairs. Blocks at the same locations of two U-Nets have shortcut connections. Figure \ref{fig:coupled-unet} gives an illustration. 

The coupled U-Nets still have a main feature flow along the U-Nets cascade. Let $m$ denote the feature number in the main flow and $n$ represent the generated feature number at each block of U-Net. For each block of the $i^{th}(i\geq 0)$ U-Net, its inputs contain the $m$ features in the main flow and another $n\times i$ features from the shortcut connections of previous U-Nets. They are concatenated channel-wise to $m+n\times i$ features. Then a $1\times 1$ convolution compresses them to $4\times m$ features. A following $3\times 3$ convolution produces $n$ new features. At last, the $m+n\times i$ input features and $n$ generated features are concatenated. Another $1\times 1$ convolution compresses them to $m$ output features, flowing into the next block.

Intuitively, the coupled U-Nets are stacked U-Nets plus the shortcut connections among the semantic blocks. Therefore, the coupled U-Nets still possess the two advantages of stacked U-Nets: multiple stages top-down and bottom-up inference and the effective intermediate supervisions. Moreover, the additional shortcut connections largely boost the information flow across U-Nets.

The proposed coupling helps not only feature reuse but also the gradient backpropagation. The intermediate supervisions are known to provide additional gradients. Hence, they have an overlapping function. It is interesting to investigate how they cooperate with each other. Empirically, coupled U-Nets with moderate intermediate supervisions would achieve the highest prediction accuracy. However, the stacked U-Nets usually work the best with full intermediate supervisions. The coupling makes some intermediate supervisions not necessary.

\section{Experiments}
In the experiments, we apply the CU-Net on the human pose estimation. First, we compare different hyper-parameter configurations of the CU-Net and choose one setting with the trade-off of accuracy and parameter efficiency. Then we investigate how the CU-Net performs with intermediate supervisions. After that, we compare the CU-Net with the naive dense U-Net. At last, we compare the CU-Net with state-of-the-art human pose estimators in terms of both accuracy and the parameter number.

{\bf Training.} We implement the CU-Net based on the Pytorch toolbox. The optimizer RMSprop is used to train the networks. The initial learning rate starts from $2.5\times 10^{-4}$ which is decayed to $5\times 10^{-5}$ after the validation accuracy becomes stable.

{\bf Datasets.} For human pose estimation, we use benchmark datasets: MPII Human Pose \cite{andriluka14cvpr} and Leeds Sports Pose (LSP) \cite{johnson2010lsp}. We also use random scaling (0.75-1.25), rotation (-/+30) and left-right flip to augment the data. We measure the human pose estimation accuracy by the Percentage of Correct Keypoints (PCK). More specifically, we PCKh@0.5 and PCK@0.2 are used to measure the accuracy on MPII and LSP respectively.

\subsection{Hyper-Parameter Selection}
\begin{figure}[th]
\minipage[t]{0.49\textwidth}
\centering
  \includegraphics[width=\linewidth]{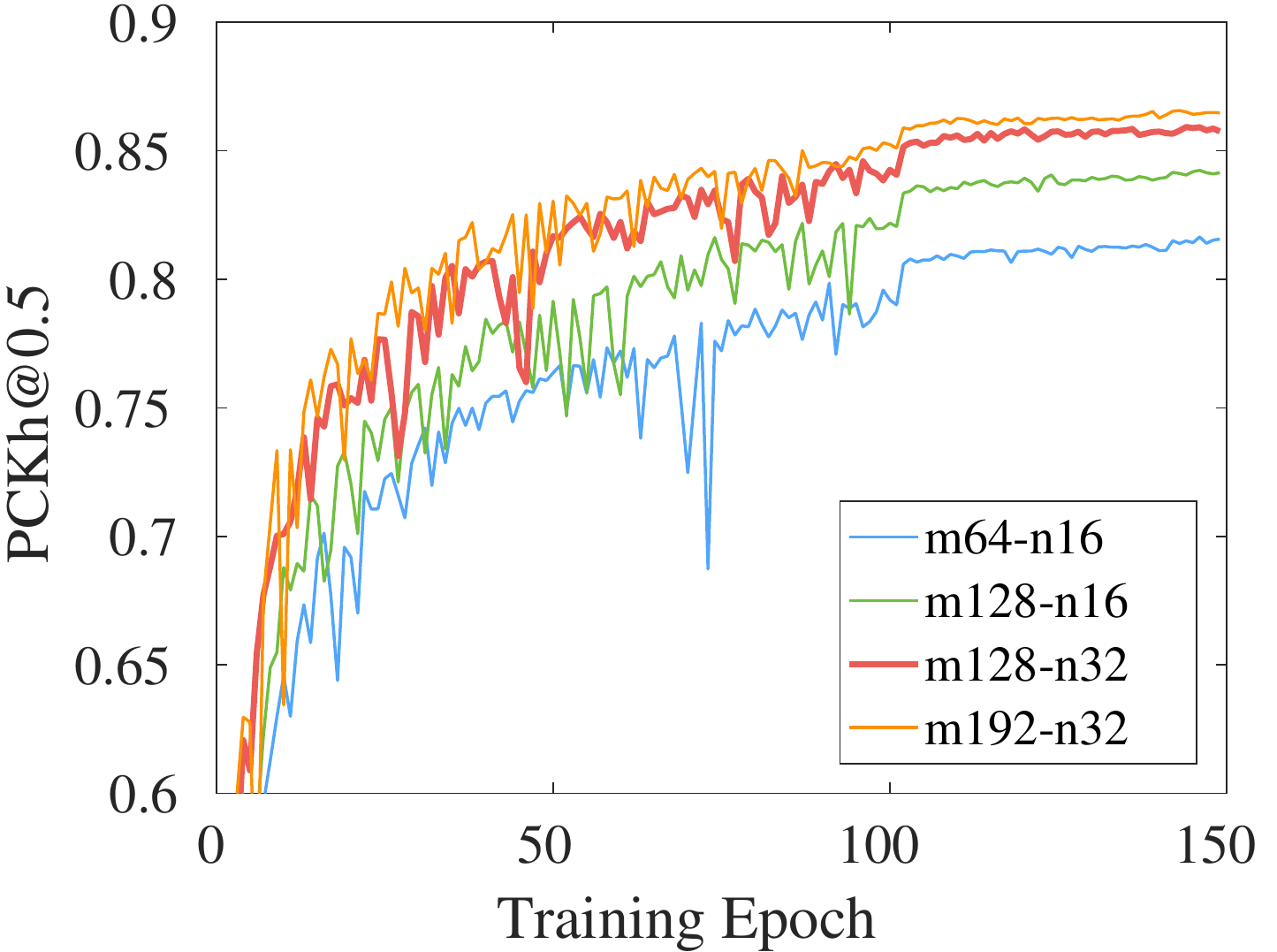}
  \caption{Curves of validation PCKh under different hyper-parameters $m$ and $n$. The converged curve reaches higher for larger $m$ and $n$. But the gap between adjacent curves becomes smaller. Larger $m$ and $n$ also make the curve smoother, indicating more stable training.}
  \label{fig:mn-para}
\endminipage \hfill
\minipage[t]{0.49\textwidth}
\centering
  \includegraphics[width=\linewidth]{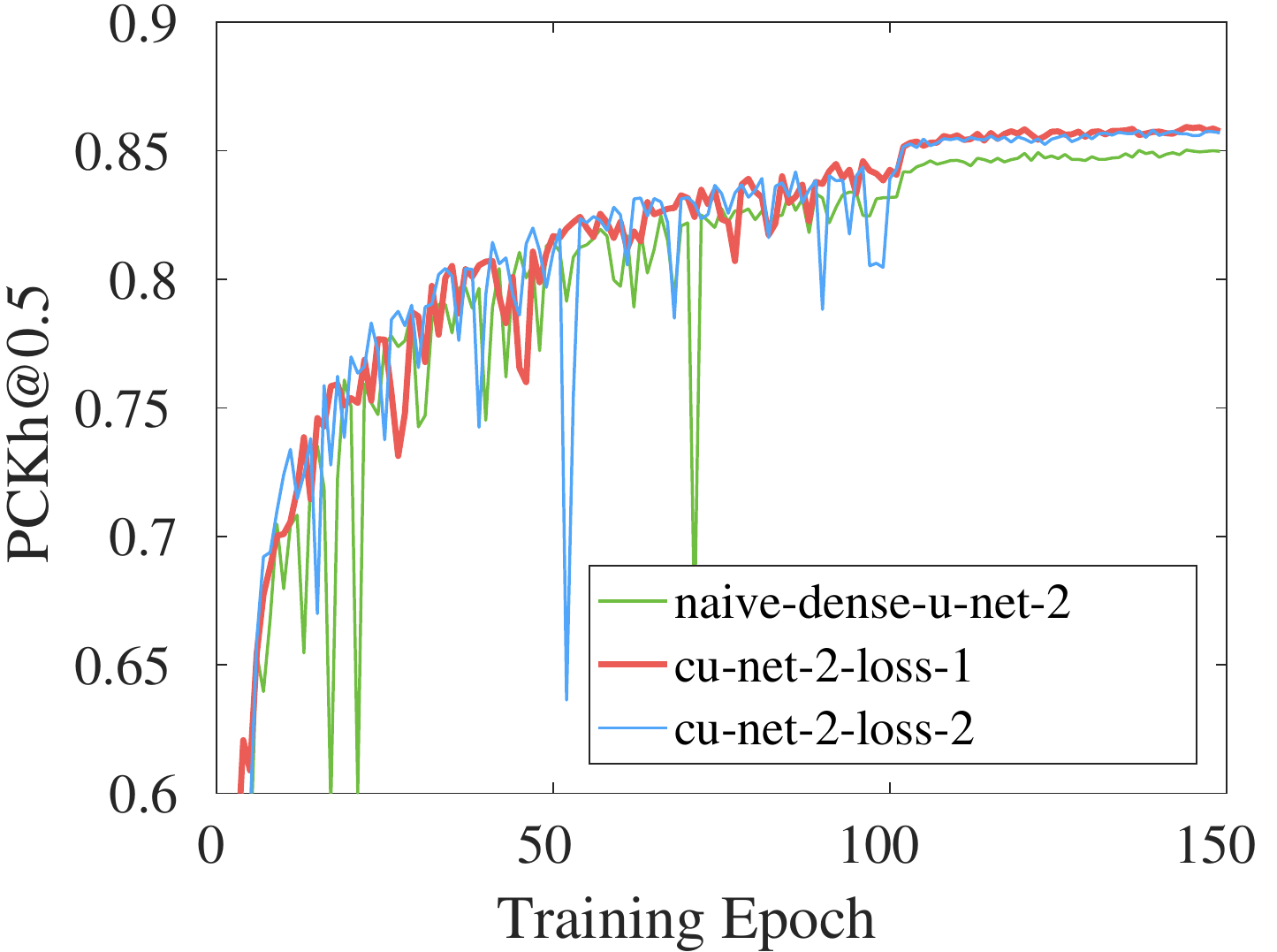}
  \caption{Validation PCKh curves of a dense U-Net, 2 coupled U-Nets (CU-Net-2) with 1 and 2 supervisions. The CU-Net-2 outperforms the dense U-Net. The CU-Net-2 with 2 supervisions does not further improve the PCKh because CU-Net-2 is not deep enough.}
  \label{fig:cu-net-2-vs-naive} \hfill
\endminipage
\vspace{-10pt}
\end{figure}

There are two important hyper-parameters in designing the CU-Net. One is the feature number $m$ in the main feature stream. In the experiments, $m$ remains the same when the feature map resolution changes. The other hyper-parameter is the generated feature number $n$ in a block of U-Net. We have tried 6 combinations of $m$ and $n$. Table \ref{tb:mn-para} gives the PCKhs on the MPII validation set. Besides, we choose 4 from the 6 settings and show how their validation PCKhs change during the training process in Figure \ref{fig:mn-para}.

In Table \ref{tb:mn-para}, the smallest $m$ and $n$ are 64 and 16. We set the increments 64 and 8 for $m$ and $n$. We could observe how the accuracy (PCKh) and the parameter number change along with the two hyper-parameters. First, the accuracy increases when $m$ and $n$ grow. Furthermore, the increase is 2.6\%, 1.4\%, 0.4\%, 0.3\% and 0.3\% from the left to the right. The increase slows down. Similar phenomena could be observed in Figure \ref{fig:mn-para}. The training is more stable when $m$ and $n$ become larger according to the curves in Figure \ref{fig:mn-para}.

Besides, the parameter number also grows as $m$ and $n$ become larger. Moreover, the growths are 0.5M, 0.4M, 0.5M, 0.5M and 0.5M. The parameter growth remains consistent. We would like to select a model with high accuracy and low model complexity. Through balancing the accuracy and parameter number, we choose $m$=128 and $n$=32. We fix this setting in the following experiments.


\begin{table}[t!]
\begin{center}
\caption{Comparison of different hyper-parameters $m$ and $n$ measured by the model parameter number and the PCKh on the MPII validation set. The PCKh increase becomes less from the left to the right while the parameter number growly consistently. A good trade-off between the PCKh and parameter number is $m$=128 and $n$=32.}\label{tb:mn-para}
\begin{tabular}{c|cccccc}
\toprule
$m$ & 64 & 128 & 128 & 128 & 192 & 192\\
\hline
$n$ & 16 & 16 & 24 & 32 & 24 & 32\\
\hline
\# Parameters & 0.5M & 1.0M & 1.4M & 1.9M & 2.4M  & 2.9M\\
\hline
PCKh@0.5 (\%) & 81.6 & 84.2 & 85.6 & 86.0 & 86.3 & 86.6\\
\bottomrule
\end{tabular}  
\end{center}
\end{table}

\subsection{Investigation of CU-Net with Intermediate Supervisions}
Generally, the supervision of a CU-Net is the supervision of its last U-Net. Since a CU-Net  contains several U-Nets, we consider to add supervisions for preceding U-Nets. More specifically, we only add the supervision at the end of a U-Net. Fortunately, the coupling connections do not prevent us from doing this. Note that if the supervision number is smaller than the U-Net number, we distribute the supervisions as uniformly as possible. For example, if 2 supervisions exist in 4 coupled U-Nets, they are at the end of the second and fourth U-Nets. 

Table \ref{tab:inter-loss} gives the PCKh comparison of CU-Net with different number of supervisions. For 2 coupled U-Nets, adding a supervision for the first U-Net makes the validation PCKh drop by 0.2\%. The coupling connections already strengthen the gradient propagation. The additional supervision makes the gradient too strong so that the model overfits the training set a little bit.

However, observations are different for more coupled U-Nets. According to Table \ref{tab:inter-loss}, additional supervisions could improve the PCKh of 4 coupled U-Nets (CU-Net-4). However, the CU-Net-4 obtains the highest PCKh with 1 additional supervision. Similar results appear for the CU-Net-8. But 3 additional supervisions help get the highest PCKh. CU-Net-4 and CU-Net-8 are much deeper than the CU-Net-2. The coupling connections still could not compensate the gradient vanishing due to the long distance propagation. Thus, adding some intermediate supervisions could further improve the accuracy. The CU-Net-8 is twice deeper than the CU-Net-4, thereby requiring more additional intermediate supervisions.

\begin{table*}[t!]
\centering
\caption{PCKhs of the CU-Net with varied intermediate supervisions on the MPII validation set. CU-Net-2 denotes a CU-Net with 2 U-Nets. The intermediate supervisions lower the PCKh of CU-Net-2. However, it improves the PCKh of deeper networks CU-Net-4 and CU-Net-8. Deeper CU-Net requires more intermediate supervisions to get the highest PCKh. But full intermediate supervisions are not the optimal.}
\label{tab:inter-loss}.
\setlength\tabcolsep{5pt}
\begin{tabular}{l c c c c c c c c c c c c}
\toprule
& \multicolumn{2}{c}{CU-Net-2} & & \multicolumn{4}{c}{CU-Net-4} && \multicolumn{4}{c}{CU-Net-8}\\
\cline{2-3} \cline{5-8} \cline{10-13} 
\# Supervisions & 1 & 2 & & 1 & 2 & 3 & 4 & & 1 & 2 & 4 & 8 \\
\cline{2-3} \cline{5-8} \cline{10-13} 
PCKh@0.5 (\%) & 86.0 & 85.8 && 87.6 & 88.1 & 88.0 & 87.8 && 88.6 & 89.3 & 89.4 & 89.0\\
\bottomrule
\end{tabular}
\end{table*}

\subsection{Comparison of CU-Net with Naive Dense U-Net}
\begin{figure*}[t!]
\centering
  \includegraphics[width=.9\linewidth]{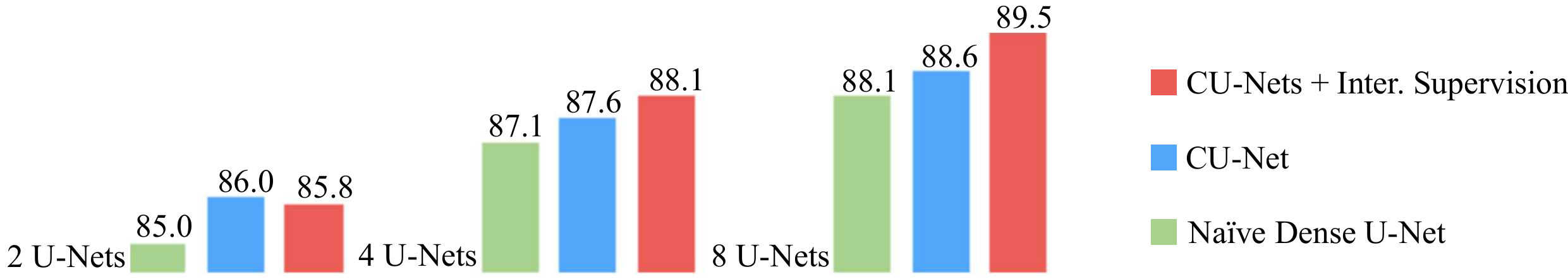}
\caption{CU-Net {\it v.s.} naive dense U-Net measured by the PCKh on the MPII validation set. In the three comparisons, the CU-Net has 2, 4 and 8 U-Nets. The naive dense U-Net is a single U-Net with equivalent sizes. The CU-Net outperforms the dense U-Net obviously. Adding intermediate supervisions could help further improve the PCKh for deep CU-Net.}
\label{fig:cu-net-vs-naive-hist}
\end{figure*}

We design the CU-Net after analyzing the drawbacks of the naive dense U-Net. In this experiment, we compare them to validate the design. The overall PCKh comparison of naive dense U-Net, CU-Net and CU-Net with intermediate supervisions are shown in Figure \ref{fig:cu-net-vs-naive-hist}. It shows three groups of comparisons with 2, 4 and 8 U-Nets. Note that the dense U-Net is always a single U-Net. For fair comparison, we add one layer in each dense block of the dense U-Net every time we increase one U-Net in the CU-Net. 

According to Figure \ref{fig:cu-net-vs-naive-hist}, the CU-Net obviously outperforms the dense U-Net by 1.0\%, 0.5\% and 0.5\% from the left to the right. This demonstrates the multi-stage top-down bottom-up inference in the CU-Net could improve the accuracy. And the CU-Net and dense U-Net have the same number of parameters in the three settings. Further, adding intermediate supervisions could improve the accuracy except for the 2 U-Nets setting. Because larger networks requires more supervisions to help the training. This proves that the CU-Net has the flexibility to use intermediate supervisions. It is worth pointing out that both the repeated top-down and bottom-up processing and the intermediate supervisions do not require extra parameters.

\begin{figure}[th]
\minipage[t]{0.49\textwidth}
\centering
  \includegraphics[width=\linewidth]{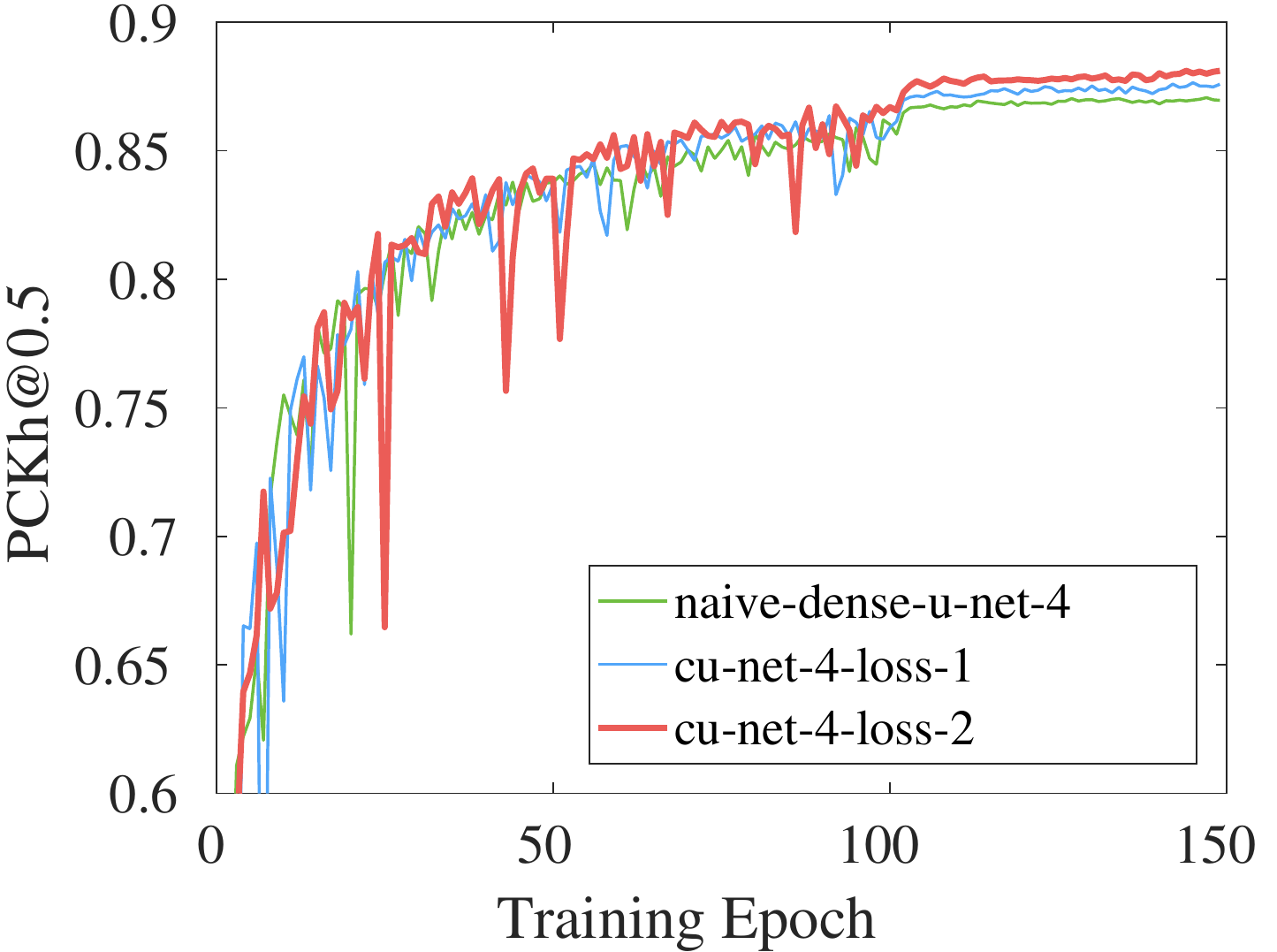}
  \caption{Validation PCKh curves of a dense U-Net and 4 coupled U-Nets (CU-Net-4) with 1 and 2 supervisions. There are clear gaps between the three converged curves. The intermediate supervision makes the curve fluctuate more from about 30 to 90 training epochs.}
\label{fig:cu-net-4-vs-naive}
\endminipage \hfill
\minipage[t]{0.49\textwidth}
\centering
  \includegraphics[width=\linewidth]{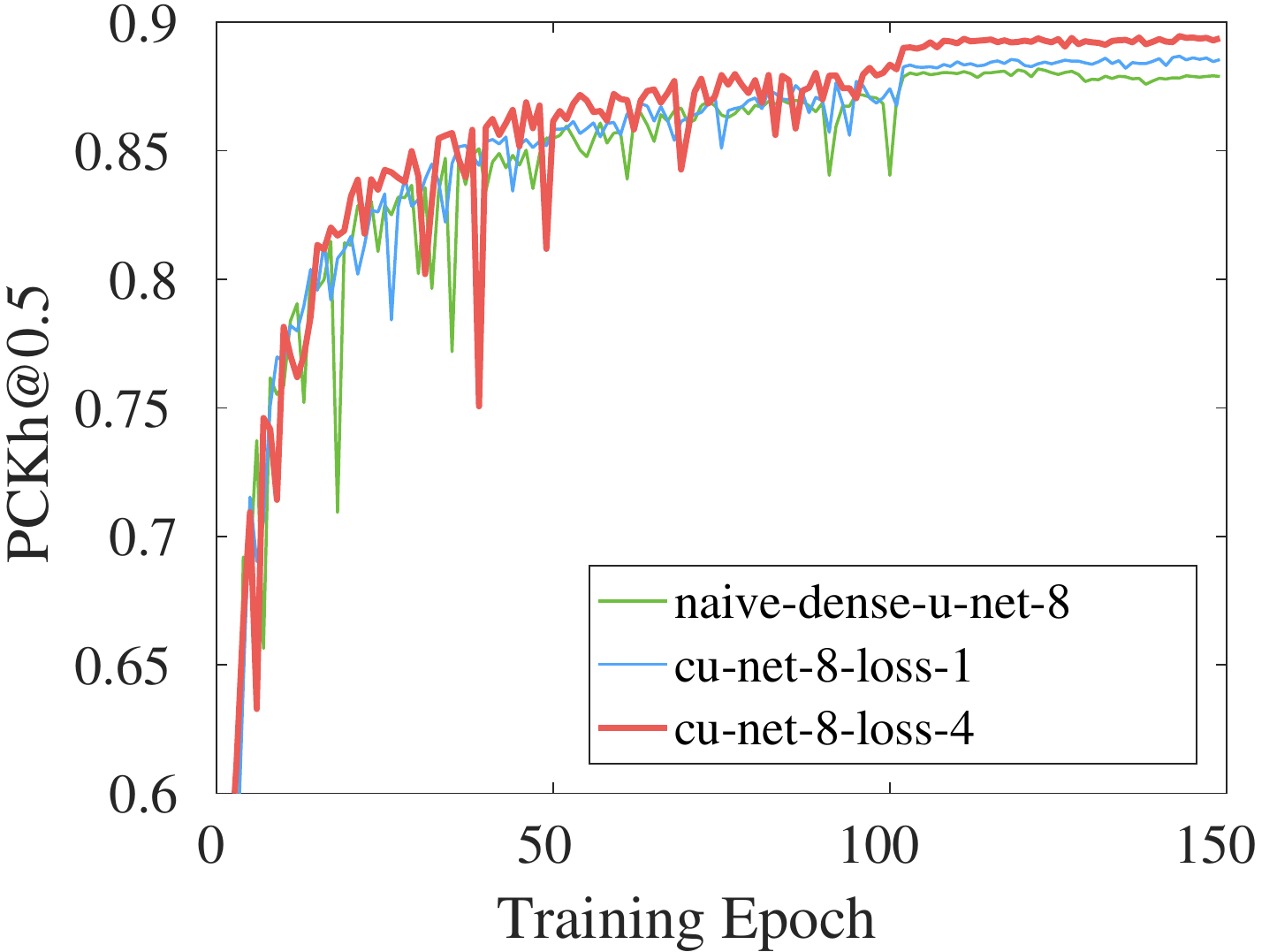}
  \caption{Validation PCKh curves of a dense U-Net and 8 coupled U-Nets (CU-Net-8) with 1 and 4 supervisions. CU-Net-8 and CU-Net-4 with 3 and 1 intermediate supervisions gets the highest PCKhs. Deeper CU-Net usually benefits from more intermediate supervisions. }
  \label{fig:cu-net-8-vs-naive} \hfill
\endminipage
\vspace{-10pt}
\end{figure}

We also show their PCKh curves under the three settings in Figures \ref{fig:cu-net-2-vs-naive}, \ref{fig:cu-net-4-vs-naive} and \ref{fig:cu-net-8-vs-naive}. The converged PCKh gaps are consistent with those in Figure \ref{fig:cu-net-vs-naive-hist}. Besides, the PCKh curve fluctuates more when adding the intermediate supervisions. The model learner would make larger steps on the training set with more supervisions. Due to the distribution shift of training and validation sets, it is easier to over-shoot the local minimas in the validation set.

\subsection{Comparison with State-of-the-art Methods}

In this experiment, we compare 8 coupled U-Nets (CU-Net-8) with state-of-the-art approaches for human pose estimation. Based on above experiments, we choose hyper-parameters $m=128$ and $n=32$ and use intermediate supervisions with the CU-Net. More specifically, we use supervisions for the $2^{nd}$, $4^{th}$, $6^{th}$ and $8^{th}$ U-Nets.

Table \ref{tb:mpii-lsp} shows comparisons of human pose estimation on MPII and LSP test sets. The  CU-Net-8 achieves comparable PCKhs as state-of-the-art methods. In contrast, as shown in Table \ref{tb:para-num}, the CU-Net-8 has only 17\%-40\% parameters of other recent state-of-the-art methods. It is worth highlighting that Newell {\it et al.} \cite{newell2016stacked} use 8 stacked U-Nets. The CU-Net-8 could obtain comparable PCKhs but with only 40\% parameters.

The CU-Net is simple and effective. Other state-of-the-art methods use stacked U-Nets with either sophisticated modules \cite{yang2017learning}, graphical models \cite{chu2017multi} or extra adversarial networks \cite{yu2017adversarial}.



\begin{table}[t]
\begin{center}
\small
\caption{Comparison of model parameter numbers with state-of-the-art human pose estimators. The CU-Net-8 uses only 17\%-40\% parameters of other methods. In particular, the CU-Net-8 has 40\% parameters of 8 stacked U-Nets \cite{newell2016stacked}. The coupling connections makes the CU-Net much more parameter efficient.}\label{tb:para-num}
\setlength\tabcolsep{3pt}
\begin{tabular}{lcccccccc}
\toprule
\multirow{2}{*}{Method} & Yang  & Wei & Bulat  & Chu & Newell  & CU-Net\\
& {\it et al.}\cite{yang2017learning} & {\it et al.}\cite{wei2016convolutional}  
& {\it et al.}\cite{bulat2016human} & {\it et al.}\cite{chu2017multi} & {\it et al.}\cite{newell2016stacked} & (8 U-Nets)\\
\hline
\# Parameters & 28.0M & 29.7M & 58.1M & 58.1M & 25.5M & {\bf 10.1M}\\
\bottomrule
\end{tabular}
\end{center}
\vspace{-10pt}
\end{table}

\begin{table}[t!]
\begin{center}
\small
\setlength\tabcolsep{1.5pt}
\caption{PCKh comparison on MPII ({\bf Top}) and LSP ({\bf Bottom}) test sets. The CU-Net-8 could achieve comparable performance as state-of-the-art methods. More importantly, it is concise and simple. In contrast, recent state-of-the-arts approaches using more sophisticated modules, graphical models or additional adversarial networks.}\label{tb:mpii-lsp}
\begin{tabular}{@{}lcccccccc@{}}
\toprule
Method & Head & Sho. & Elb. & Wri. & Hip & Knee & Ank. & Mean\\
\hline
Pishchulin \textit{et al.} ICCV'13 \cite{pishchulin2013strong} & 74.3 & 49.0 & 40.8 & 34.1 & 36.5 & 34.4 & 35.2 & 44.1\\
Tompson \textit{et al. } NIPS'14 \cite{tompson2014joint} & 95.8 & 90.3 & 80.5 & 74.3 & 77.6 & 69.7 & 62.8 & 79.6\\
Carreira \textit{et al.} CVPR'16 \cite{carreira2016human} & 95.7 & 91.7 & 81.7 & 72.4 & 82.8 & 73.2 & 66.4 & 81.3\\
Tompson \textit{et al.} CVPR'15 \cite{tompson2015efficient}& 96.1 & 91.9 & 83.9 & 77.8 & 80.9 & 72.3 & 64.8 & 82.0\\
Hu \textit{et al.} CVPR'16 \cite{hu2016bottom}& 95.0 & 91.6 & 83.0 & 76.6 & 81.9 & 74.5 & 69.5 & 82.4\\
Pishchulin \textit{et al.} CVPR'16 \cite{pishchulin2016deepcut}&94.1 & 90.2 & 83.4 & 77.3 & 82.6 & 75.7 & 68.6 & 82.4\\
Lifshitz \textit{et al.} ECCV'16 \cite{lifshitz2016human} & 97.8 & 93.3 & 85.7 & 80.4 & 85.3 & 76.6 & 70.2 & 85.0\\
Gkioxary \textit{et al.} ECCV'16 \cite{gkioxari2016chained} & 96.2 & 93.1 & 86.7 & 82.1 & 85.2 & 81.4 & 74.1 & 86.1\\
Rafi \textit{et al.} BMVC'16 \cite{rafi2016efficient} & 97.2 & 93.9 & 86.4 & 81.3 & 86.8 & 80.6 & 73.4 & 86.3\\
Belagiannis \textit{et al.} FG'17 \cite{belagiannis2017recurrent}&97.7 & 95.0 & 88.2 & 83.0 & 87.9 & 82.6 & 78.4 & 88.1\\
Insafutdinov \textit{et al.} ECCV'16 \cite{insafutdinov2016deepercut}&96.8 & 95.2 & 89.3 & 84.4 & 88.4 & 83.4 & 78.0 & 88.5\\
Wei \textit{et al.} CVPR'16 \cite{wei2016convolutional} & 97.8 & 95.0 & 88.7 & 84.0 & 88.4 & 82.8 & 79.4 & 88.5\\
Bulat \textit{et al.} ECCV'16 \cite{bulat2016human} & 97.9 & 95.1 & 89.9 & 85.3 & 89.4 & 85.7 & 81.7 & 89.7\\
Newell {\it et al.}  ECCV'16 \cite{newell2016stacked} & 98.2 & 96.3 & 91.2 & 87.1 & 90.1 & 87.4 & 83.6 & 90.9\\
Chu \textit{et al.} CVPR'17 \cite{chu2017multi} & 98.5 & {\bf 96.3} & {\bf 91.9} & {\bf 88.1} & {\bf 90.6} & {\bf 88.0} & {\bf 85.0} & {\bf 91.5}\\
\hline
CU-Net-8 & 97.4 & 96.2  & 91.8  & 87.3  & 90.0  & 87.0 & 83.3 & 90.8\\
\bottomrule
\toprule
Method & Head & Sho. & Elb. & Wri. & Hip & Knee & Ank. & Mean\\
\hline
Belagiannis \textit{et al.} FG'17 \cite{belagiannis2017recurrent} & 95.2 & 89.0 & 81.5 & 77.0 & 83.7 & 87.0 & 82.8 & 85.2\\
Lifshitz \textit{et al.} ECCV'16 \cite{lifshitz2016human} & 96.8 & 89.0 & 82.7 & 79.1 & 90.9 & 86.0 & 82.5 & 86.7\\
Pishchulin \textit{et al.} CVPR'16 \cite{pishchulin2016deepcut} &  97.0 & 91.0 & 83.8 & 78.1 & 91.0 & 86.7 & 82.0 & 87.1\\
Insafutdinov \textit{et al.} ECCV'16 \cite{insafutdinov2016deepercut}& 97.4 & 92.7 & 87.5 & 84.4 & 91.5 & 89.9 & 87.2 & 90.1\\
Wei \textit{et al.} CVPR'16 \cite{wei2016convolutional}& 97.8 & 92.5 & 87.0 & 83.9 & 91.5 & 90.8 & 89.9 & 90.5\\
Bulat \textit{et al.} ECCV'16 \cite{bulat2016human}& 97.2 & 92.1 & 88.1 & 85.2 & 92.2 & 91.4 & 88.7 & 90.7\\
Chu \textit{et al.} CVPR'17 \cite{chu2017multi}& 98.1 & 93.7 & 89.3 & 86.9 &  93.4 & 94.0 & 92.5 & 92.6\\
Newell {\it et al.} ECCV'16 \cite{newell2016stacked} & {\bf 98.2} & 94.0 & 91.2 & 87.2 & 93.5 & {\bf 94.5} & 92.6 & 93.0\\
\hline
CU-Net-8 & 97.1 &  {\bf 94.7} & {\bf 91.6} & {\bf 89.0}  & {\bf 93.7} & 94.2 & {\bf 93.7}  & {\bf 93.4}\\
\bottomrule
\end{tabular}
\end{center}
\vspace{-10pt}
\end{table}
\section{Conclusion}
We have proposed the CU-Net, a new architecture based on the U-Net. We connect the same semantic blocks of several stacked U-Nets. Each U-Net pair are coupled since they are connected at multiple resolutions. Compared with the naive dense U-Net, the CU-Net has the advantages of multi-stage top-down and bottom-up inference and intermediate supervisions. Compared with the stacked U-Nets, it is more parameter efficient benefiting from the feature reuse across U-Nets. Experiments on MPII and LSP benchmark datasets show that it could achieve state-of-the-art accuracy but using at most 40\% model parameters of other methods.
\section{Acknowledgment}
This work is partly supported by the Air Force Office of Scientific Research (AFOSR) under the Dynamic Data-Driven Application Systems Program, NSF 1763523, 1747778, 1733843 and 1703883 Awards.
\bibliography{bmvc18.bib}
\end{document}